\documentclass{article}
\usepackage{spconf,amsmath,graphicx}

\usepackage[noadjust]{cite}

\usepackage[utf8]{inputenc}
\usepackage{booktabs}
\usepackage{multicol}
\usepackage{graphicx}
\usepackage{multirow}
\usepackage{amssymb,amsmath,bm}
\usepackage{enumitem}
\usepackage{soul,color}
\usepackage{array}
\usepackage{footnote}
\makesavenoteenv{table}
\usepackage{hyperref}
\usepackage{color}
\usepackage{graphicx}
\usepackage{subfigure}
\usepackage{xcolor}
\PassOptionsToPackage{hyphens}{url}\usepackage{hyperref}

\usepackage[left]{lineno}
\usepackage[utf8]{inputenc}
\usepackage{amsmath,bm}
\usepackage[ruled, linesnumbered]{algorithm2e}
\newcommand{\bs}[1]{\boldsymbol{-1}} 
\newcommand{\addcomment}[1]{\hl{\textbf{NOTE:}}\hl{\emph{-1}}}

\usepackage{float}
\title{SALM: Speech-augmented Language Model with In-context Learning for Speech Recognition and  Translation}
\name{
\begin{tabular}{c} 
Zhehuai Chen$^*$, He Huang$^*$, Andrei Andrusenko, Oleksii Hrinchuk, \\
Krishna C. Puvvada, Jason Li, Subhankar Ghosh, Jagadeesh Balam, Boris Ginsburg\thanks{Thanks to Aleksandr Laptev, Somshubra Majumdar, Nithin Koluguri, Paarth Neekhara, Xuesong Yang, Vitaly Lavrukhin, Rafael Valle, Yi Dong, Adi Renduchintala, Sandeep Subramanian, Yang Zhang for discussion.}
\end{tabular}
}
\address{NVIDIA, USA}
\begin{document}
\ninept
\maketitle
\begin{abstract}

We present a novel Speech Augmented Language Model (SALM) with {\em multitask} and {\em in-context} learning capabilities. SALM comprises a frozen text LLM, a audio encoder, a modality adapter module, and LoRA layers to accommodate speech input and associated task instructions. The unified SALM not only achieves performance on par with task-specific Conformer baselines for Automatic Speech Recognition (ASR) and Speech Translation (AST), but also exhibits zero-shot in-context learning capabilities, demonstrated through keyword-boosting task for ASR and AST. 
Moreover,  {\em speech supervised in-context training} is proposed to bridge the gap between LLM training and downstream speech tasks, which  further boosts the in-context learning  ability of speech-to-text models.
Proposed model is open-sourced via  NeMo toolkit~\footnote{\scriptsize\url{https://github.com/NVIDIA/NeMo/tree/modular_speechllm}}.

\end{abstract}

\def\thefootnote{*}\footnotetext{Equal contribution}

\vspace{-1em}
\section{Introduction}
\label{sec:intro}
\vspace{-0.4em}

Large language models (LLMs) have achieved remarkable results on a variety of natural language processing (NLP) benchmarks recently~\cite{openai2023gpt,anil2023palm}. These models can be trained on massive amounts of unsupervised text data, and learn the knowledge that benefits many downstream text generative tasks. Through instruction tuning, LLMs can be fine-tuned to be more amenable to solving different NLP tasks in general. Additionally, LLMs demonstrate an in-context learning ability, meaning that they can learn from a few examples in the context, even if those examples are unseen in the training data.

These properties of LLMs are attractive to other modalities, including speech. Different interfaces between speech and LLMs have been studied, including text~\cite{huang2023audiogpt, chen2023x,chen2023large,ma2023n}, 
quantized audio tokens~\cite{zhang2023speechgpt,rubenstein2023audiopalm} and continuous audio embeddings~\cite{gong2023listen,ling2023adapting,wu2023decoder,fathullah2023prompting,wang2023speech}. Promising results have been shown in speech recognition, translation and synthesis. 

In this work,  we prompt Megatron LLM\cite{shoeybi2019megatron} using NeMo\cite{kuchaiev2019nemo} speech models with different motivations: i)  utilize the multitask ability of LLMs to construct a unified model for various speech tasks.
ii) augment speech models with the in-context learning (ICL) ability of LLMs.
 Our main contributions include:

\begin{itemize}[leftmargin=*]
    \item  Propose SALM which performs multitask speech-language modeling in a unified LLM framework. The unified model performs on par with bespoke Conformer baselines in  ASR and AST. The speech-LLM solution is open-sourced via  NeMo \cite{kuchaiev2019nemo}.
    \item Equip speech-to-text models with zero-shot in-context learning ability for the first time, shown by   ASR and AST keyword boost. 
    \item Propose {\em speech supervised in-context training} to further boost  ICL ability of speech models.
\end{itemize}

\vspace{-1em}
\section{Related Work}
\label{sec:related}
\vspace{-0.4em}
The success of LLMs in NLP tasks~\cite{openai2023gpt,anil2023palm}, has motivated growing interest in leveraging them to improve speech modeling.
This work focuses on speech-to-text applications.
One set
of approaches use text as the interface between speech models and LLMs\cite{huang2023audiogpt, chen2023x,chen2023large,ma2023n}. 
Recently \cite{sun2023can} looks into using GPT-2 in the N-best rescoring 
 for contextual ASR.
However,
some information in the speech modality may be lost due to the hardness in capturing them through text, e.g. speaker information, emotion and accents. In contrast, recent research starts to look at deep integration between speech models and LLMs, e.g. SpeechGPT~\cite{zhang2023speechgpt}, AudioPaLM~\cite{rubenstein2023audiopalm}, LTU~\cite{gong2023listen}, etc.~\cite{ling2023adapting,wu2023decoder,fathullah2023prompting,wang2023speech}.
Among them, Speech-LLaMA~\cite{wu2023decoder,fathullah2023prompting} are the most relevant to this work, which share an architecture of prepending  continuous audio embeddings to the text  embeddings before feeding to a decoder-only LLM. 

This work advances the previous works by equipping speech-to-text models with in-context learning (ICL) ability,
demonstrated by keyword boosting tasks in ASR and AST.
Extending ICL to speech domain is under-explored. Previous works VALL-E\cite{wang2023neural} and Voicebox\cite{le2023voicebox} focus on the text-to-speech models. 
Moreover, we will open-source our implementation to accelerate this line of research.

The keyword boosting and contextual speech recognition have been explored in previous speech models. 
One branch of methods use external keyword LMs and WFSTs to bias the speech model in the inference time~\cite{williams2018contextual}. The other branch tries to integrate contextual information into the
E2E modeling (CLAS)\cite{pundak2018deep}.
%
 %
This work studies keyword boosting for speech applications with the in-context learning ability of LLMs and compares it to the first branch of methods. The proposed method does not require external context biasing graphs or learning explicit  model weights for boosted words.

Text injection is another way for speech models to benefit from text. \cite{thomas2022integrating,chen2022maestro,zhang2022speechlm} modify the speech  models to take both speech and text. \cite{zhang2023google, barrault2023seamlessm4t} scale these up and achieve remarkable success. 

\begin{figure*}[hbt!]
  \centering
    \includegraphics[width=0.7\linewidth]{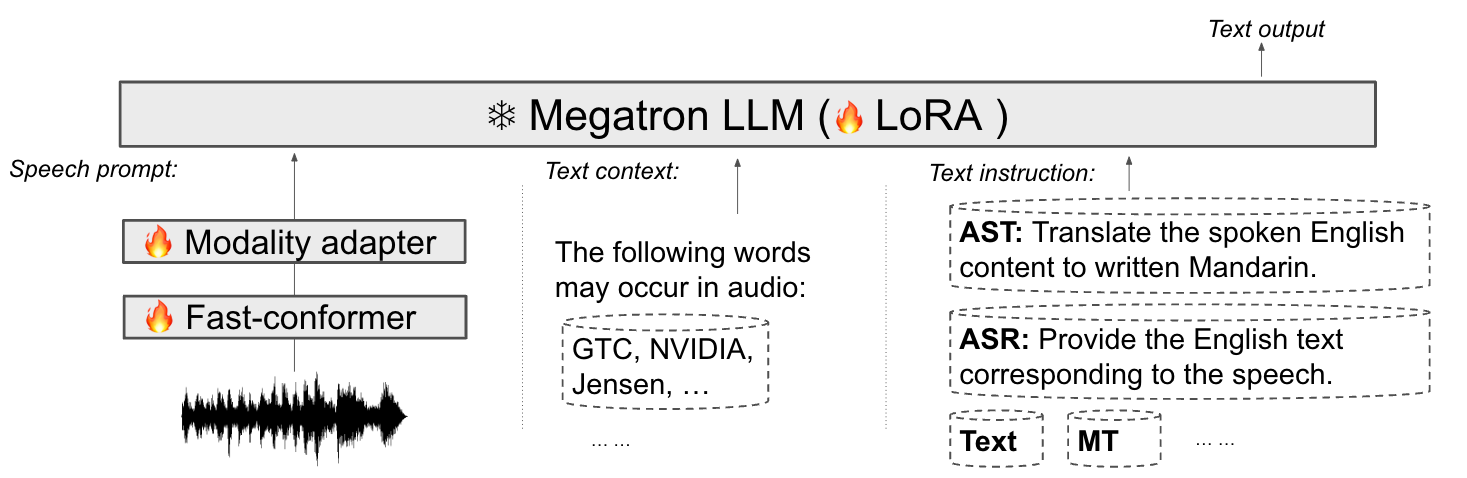}
    \vspace{-1em}
    \caption{SALM for Multitask Modeling and In-Context Learning.}
    \label{fig:framework}
    \vspace{-1.5em}
\end{figure*}

\vspace{-0.6em}
\section{SALM - Speech Augmented Language Model}
\label{sec:proposed}
\vspace{-0.4em}

This work proposes to conduct supervised speech instruction tuning directly on a text pretrained and instruction fine-tuned LLM. The resultant {\em SALM}  learns to condition on speech prompt, text context and instruction to predict textual outputs for different speech tasks, as shown in Figure~\ref{fig:framework}. 
The introduced LLM potentially equips speech-to-text models with in-context learning ability.

\subsection{Speech and text prompts}
\vspace{-0.4em}
We choose Fast Conformer~\cite{rekesh2023fast} and GPT-style Megatron LLM\cite{shoeybi2019megatron} as the speech and text backbones.
Fast Conformer is a carefully redesigned Conformer~\cite{gulati2020conformer} with a new downsampling schema for better efficiency while preserving state-of-the-art accuracy. We use a 110M pretrained audio encoder from NeMo and a pretrained 2B Megatron LLM with text instruction fine-tuned.

To guide LLM to condition on outputs from the audio encoder, we introduce modality adapter and LoRA layers~\cite{hu2021lora} described below and train these layers through {\em multitask speech instruction tuning} in Section~\ref{sec:speech-sft}. Two Conformer layers with 4X subsampling are used as modality adapter layers in this paper to match the different information rate and modeling space between text and speech. 
The resultant speech prompt has a frame-shift of 320ms. It is projected to the LLM dimension and prepended to text context and instruction as the input of LLM, shown in Figure~\ref{fig:framework}.
Low-rank Adaptation (LoRA) layers with 128 dimensions are added to LLM during the speech instruction tuning. We freeze the LLM and back-propagate the rest.

\vspace{-0.8em}
\subsection{Multitask supervised speech instruction tuning} 
\label{sec:speech-sft}
\vspace{-0.4em}
One of the central motivation of combining speech model and LLM in this work is to bring the instruction tuning~\cite{flandataset} from NLP to speech multitask learning and provide a unified speech model. 
We include different speech tasks (ASR, AST and more) with diverse instructions so as to not only promote instruction following but also improve generalization of the aforementioned  modality adapter layers on different tasks.
This work reuses paired speech and text data from ASR and AST public corpora and randomly prepends task instruction as examples in Figure~\ref{fig:framework} in the training time. 

\vspace{-0.8em}
\subsection{In-context learning for speech-to-text tasks}
\label{sec:icl-wb}
\vspace{-0.4em}

The other main motivation of SALM is to leverage the in-context learning (ICL) ability of LLM in speech tasks. 
%
ICL is 
one of the breakthrough from LLMs, to predict labels for unseen inputs without additional parameter updates. This ability was extended to the speech domain with previous works focusing on the text-to-speech (TTS) application. \cite{wang2023neural} proposed a neural codec language model that can synthesize speech for unseen speakers without fine-tuning.

\begin{figure}[]
  \centering
    \includegraphics[width=0.9\linewidth]{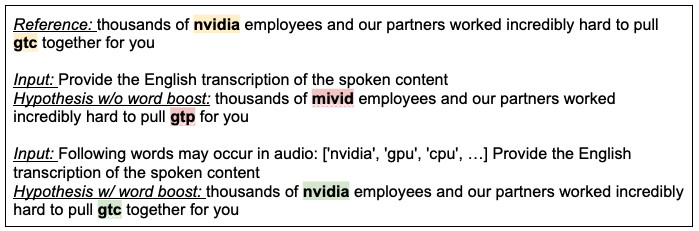}
    \vspace{-1em}
    \caption{Example of In-context Learning for Keyword Boosting.}
    \vspace{-1em}
    \label{fig:wb}
\end{figure}

In this work, we try to assess the {\em{in-context learning ability in speech understanding tasks}}, ASR and AST as examples, and improve upon it.
We take the keyword boosting task  as the first step towards this direction. Keyword boosting aims at biasing the model to recognize particular words of interest. 
We define the in-context learning here as: 
{\em learning the boosted words from the prompting text context, without back-propagation}. 
As demonstrated in  Figure~\ref{fig:wb}, we provide keywords to the model in the format of optional text context before  text instruction. 
As a contrast, previous keyword boosting works require either learning explicit embeddings and weights for  boosted words during training or with external biasing graphs.
%

\vspace{-0.8em}
\subsection{Speech supervised in-context training}
\label{sec:ssicl}
\vspace{-0.4em}

Given the differences in both data formats and learning criteria between LLM pretraining and ICL stages, previous NLP research suggests  
a series of supervised in-context finetuning strategies by constructing in-context training data to enhance ICL capability~\cite{dong2022survey}.
With similar motivation, the {\em{speech supervised in-context training}} (Speech ICT) is proposed in this work to promote the model to leverage the aforementioned text context in speech understanding.

\begin{figure}[]
  \centering 
    \includegraphics[width=1\linewidth]{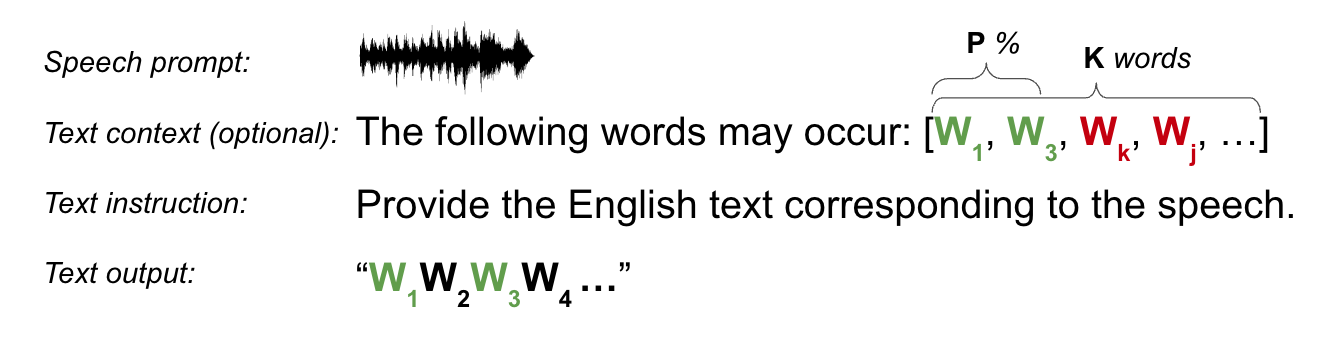}
    \vspace{-2em}
    \caption{Demonstration of the Proposed {\em Speech Supervised In-context Training}. The supervised data is augmented by including the optional text context with a probability of 5\%, where $K$ words are sampled with $P\%$ of words from the ground-truth ({\em{positive ratio}}). }
    \vspace{-1em}
    \label{fig:speechict}
\end{figure}

In the speech instruction tuning stage, we augment the same supervised data by randomly sampling words from the current utterance and other utterances in the dataset, and including them as the optional text context for the utterance as Figure~\ref{fig:speechict}. We will later demonstrate in the experiment that this way of in-context training can 
generalize to unseen words, and corpora in unseen domains.

\vspace{-0.4em}
\section{Data and Experimental Setup}
\label{sec:setup}
\vspace{-0.4em}

\textbf{Model Details}: 
The whole pipeline is implemented via  NeMo toolkit\cite{kuchaiev2019nemo}.
The audio encoder is initialized from the NGC ASR pretrained Fast Conformer-large\footnote{\scriptsize\url{https://catalog.ngc.nvidia.com/orgs/nvidia/teams/nemo/models/stt_en_fastconformer_transducer_large}}  or the Conformer self supervised learning (SSL) checkpoint\footnote{\scriptsize\url{https://catalog.ngc.nvidia.com/orgs/nvidia/teams/nemo/models/ssl_en_conformer_large}}, while the modality adapter is randomly initialized. 
%
%
The Megatron LLM \cite{shoeybi2019megatron} we used has 2B parameters, 
which was trained on 1.1T tokens on a dataset that comprises 70\% English, covering web-crawl data, news, conversation, books, and scientific domains, 
15\% Code from the Stack dataset~\cite{Kocetkov2022TheStack} and 15\% non-english text from CommonCrawl~\footnote{\scriptsize\url{https://commoncrawl.org/}}. 
This model was then finetuned on public instruction following datasets like~\cite{flandataset}.

\textbf{Hyper-parameters}: We train the model with 64 global batch size,  using Adam optimizer with learning rate 1e-4 and weight decay of 1e-3. Cosine annealing with 2000 warm-up steps is applied. Gradients are clipped to 5.0.   8 of A100 GPUs are used for training. We use greedy decoding in the inference by default while nucleus sampling ($t=0.2$, $p=0.95$, $k=50$)\cite{holtzman2019curious} is also tested.

\textbf{Speech Recognition}: We use the LibriSpeech~\cite{panayotov2015librispeech} training set to train SALM, and pick the best checkpoint beasd on the WER on dev sets, which is then evaluated on  \emph{test-clean} and \emph{test-other}. Our baseline model uses  NGC ASR pretrained Fast Conformer-large encoder and transducer decoder with 114M parameters.

\textbf{Speech Translation}: For speech translation, we use all English audio data available for the Offline Track of IWSLT 2023~\cite{iwslt:2023} paired with pseudo-generated translations to German and Japanese. Our training dataset consists of $2.7$M segments which corresponds to $4.8$K hours of audio. We used MuST-C v2 tst-COMMON~\cite{cattoni2021must} for evaluation. Our baseline model uses NGC ASR pretrained Fast Conformer-large encoder followed by $6$-layer Transformer decoder. We used $16384$k BPE encodings trained on texts in target language.

\textbf{Keyword Boosting}: 
For the keyword boosting evaluation we prepared an internal test set based on NVIDIA GTC talks data. 
The test set is forced aligned and segmented,  8 hours in total.
The main feature of such a data set is the presence of a large number of different acronyms, product names, and technical terms, which often have low recognition accuracy for ASR systems.
To build the keywords list we selected words and phrases with high occurrences in  GTC test set and low recognition accuracy for greedy decoding of baseline transducer model~\footnote{Examples: {\em NVIDIA, GPU, Omniverse, Geforce, NeMo, kubernetes,} etc.}. We include 64 keywords by default and study different numbers. For the evaluation of keywords recognition accuracy we consider precision,{\em P}, and recall, {\em R}, calculated from keywords  according to the alignment of the recognition results with the ground-truths. We also report F-score ($2*P*R/(P+R)$).
The baseline transducer model uses the shallow-fusion approach for the  boosting~\cite{williams2018contextual}. During beam search decoding, partial hypotheses are rescored according to the context biasing graph.
The implementation of the context biasing graph  was taken from Icefall 
toolkit\footnote{\scriptsize\url{https://github.com/k2-fsa/icefall/blob/master/icefall}} with context score 4. 
We use modified adaptive expansion search based on \cite{Kim2020AcceleratingRT} with beam width=5, alpha=2, and gamma=8.

\begin{table}[]
\caption{\label{tab:exp-asr} {SALM Results on ASR and AST tasks} }
\begin{tabular}{l|cc|cc}
 \toprule
                           & \multicolumn{2}{c|}{ASR WER}  & \multicolumn{2}{c}{AST BLEU} \\
    systems                      & \multicolumn{2}{c|}{LibriSpeech} & \multicolumn{2}{c}{MuST-C }  \\
                          &  clean & other & en-de & en-ja  \\
 \midrule
bespoke Fast Conf L+decoder      & 2.3        & 5.0  & 26.0 & 5.5      \\
\ \ \ \ + ASR pretrained encoder  & \textbf{1.8}        & \textbf{3.9}   & \textbf{31.0} & \textbf{14.8}      \\
\midrule
SALM (SSL pretrained)       & 2.7        & 6.1   & - & -     \\
SALM (ASR pretrained)       & 2.4        & 5.3   &  27.1 &  15.0   \\
\ \ \ \ + nucleus sampling       & \textbf{2.3}        & \textbf{4.8}  &29.6 & 16.5     \\
\midrule
ASR+AST SALM (nucleus s.)       & 2.6        & 6.1 &\textbf{30.7} & \textbf{16.8}     \\
\bottomrule
\end{tabular}
\vspace{-1.5em}
\end{table}

\begin{table}[]
\centering
\caption{\label{tab:example-asr} {Win and Loss Comparing SALM and Fast Conformer-Transducer, FC-T (errors are shown in \textcolor{red}{red}).} }
\begin{tabular}{lp{0.04\textwidth}|p{0.18\textwidth}|p{0.18\textwidth}}
\toprule
 & Type          & FC-T                                                                                                                & SALM                                                                                        \\
 \midrule
\multirow{4}{*}{Win} & rare word & ... a \textcolor{red}{kleptomania} like cousin snatcher & ... a kleptomaniac like cousin snatcher  \\
    \cmidrule{2-4}
&  seg-ment &  greenhorns \textcolor{red}{flat heads} & greenhorns flatheads  \\
\midrule
  Loss     & hallu-cinate & ah lida exclaimed fauchelevent  &  ah lidah exclaimed \textcolor{red}{shoot up the english transcription ...} \\
      \cmidrule{2-4}
& AM & rachel lake rachel lake ... & \textcolor{red}{routen leak routen leak} ...\\
\cmidrule{2-4}
& del-etion &   six hundred bishops four emperors ... three  hundred canonized  ...
&  six hundred \textcolor{red}{[del error]} canonized  ... \\
        \bottomrule
\end{tabular}
\vspace{-1.2em}
\end{table}

\section{Results and Analysis}
\label{sec:exp}
\vspace{-0.4em}

\vspace{-0.1em}
\subsection{Unified model for ASR and AST}
\label{sec:exp-asr}
\vspace{-0.4em}

Table~\ref{tab:exp-asr} shows the ASR results on LibriSpeech and AST results on MuST-C. We compare SALM with the bespoke baselines of ASR and AST in the first two rows.
ASR baseline uses FastConformer-large encoder and transducer decoder (FC-T).  
AST baseline uses transformer decoder instead and one model is trained on each language-pair as found to perform the best. 
The first row trains from-scratch and the second  uses the aforementioned NGC Fast Conformer ASR pretrained encoder.


The SALM model in the 3rd and 4th rows initializes the audio encoder from the aforementioned NGC SSL and ASR  checkpoints respectively.
The best SALM model in the fifth row with nucleus sampling in the LLM inference outperforms the from-scratch baseline but still behind the stronger baseline in the second row. 

For AST, we train  one SALM model in the fifth row to support both language pairs and use text instruction as shown by Figure~\ref{fig:framework} to switch between different pairs. 
We then further train one SALM model on both AST and ASR data to provide a unified model for both tasks in the last row. The unified model performs better than the two baselines and AST-only SALM. When operating on the ASR task, this model is  worse than ASR-only SALM.

To understand the strengths and weaknesses of LLM based SALM versus the baseline,  
Table~\ref{tab:example-asr} includes the
ASR hypothesis comparison. Although SALM suffers from hallucination and long-form deletion problems, it performs better on rare words and proper nouns. 
We found nucleus sampling can solve some of the former problems and result in better results. Further alleviating these problems will be our future work. 

\begin{table}[]
\caption{\label{tab:exp-wb-asr} {ASR Keyword Boosting Results on GTC Talk Test Set.} }
\begin{tabular}{lccc}
\toprule
Systems                               & boost  & WER   & F-score  ({\em P/R})  \\
\midrule
 \midrule                                    
Fast Conf L-Transducer +    & N             & 16.2 & 0.36 (0.96/0.22)      \\
              \ ASR pretrained encoder                         & Y             & \textbf{15.1} & \textbf{0.67 (0.87/0.55)}      \\
                                      \midrule
\multirow{2}{*}{SALM} & N             & 17.0  & 0.35 (0.94/0.21)      \\
                                & Y             & 15.8  & 0.56 (0.74/0.45)     \\
\ \ \ \ + nucleus sampling                                 & Y             & \textbf{14.9}  & \textbf{0.61 (0.66/0.57)}     \\
                                    \bottomrule
\end{tabular}
\vspace{-1.2em}
\end{table}

\vspace{-0.6em}
\subsection{Zero-shot in-context Learning for keyword boosting}
\label{sec:exp-word-boost}
\vspace{-0.4em}

We  study the zero-shot in-context learning ability of SALM by taking keyword boosting task as the proxy in Table~\ref{tab:exp-wb-asr}.
We took the  Fast Conformer-L Transduer (FC-T) initialized with NGC  ASR pretrained encoder and trained on LibriSpeech  as a strong baseline. 
Without boosting, the LibriSpeech-trained SALM performs on par (Row 3 v.s. 1).
We  prompt SALM for keyword boosting in Row 4 with the  text context described in Section~\ref{sec:icl-wb}. 
The better result from Row 3 to 4 demonstrates the effectiveness of  in-context learning  method. 
Nucleus sampling in the LLM inference can further boost the performance, results in the 5th row.
Compared to  baseline boosting, this method achieved similar relative boosting gains while  not requiring external biasing graphs\cite{williams2018contextual} as in  baseline or learning explicit biasing embedding\cite{pundak2018deep}.
%


Table~\ref{tab:exp-wb-sict} demonstrates the necessity  of the proposed Speech ICT. 
Although the LLM used in SALM has been instruction fine-tuned with text data, SALM in Row 1 without Speech ICT cannot effectively follow the prompt and obtain limited improvement. This shows the challenge of transferring textual knowledge to the speech domain in the current speech and LLM research.   
Speech ICT provides a route towards solving this problem. Including the augmented in-context training data designed in Section~\ref{sec:ssicl} significantly improves the performance. Tuning {\em positive ratio} in the table affects inference precision and recall -- the bigger it is the worse precision and better recall.
The best 6\% is used in the rest. 

\begin{table}[]
\centering
\caption{\label{tab:exp-wb-sict} {Improve In-context Learning with  Speech ICT. {\em positive ratio} is defined as the percentage of  ground-truth words in augment.}}
\begin{tabular}{r  r  c}
\toprule
\multicolumn{2}{c}{Speech ICT training setup}               & Eval with 64 words \\
\multicolumn{1}{l}{positive ratio, \%}            & \multicolumn{1}{l}{\# of keywords} & F-score ({\em P/R})      \\
\midrule
n/a                                           & 0                                  & 0.38 (0.82/0.25)   \\
\midrule
33\%                                          & 3                                  & 0.52 (0.59/0.47)   \\
33\%                                          & 64                                 & 0.52 (0.62/0.44)   \\
 \textbf{6\%}                                           & \textbf{64}                                 & 0.56 (0.74/0.45)   \\
3\%                                           & 64                                 & 0.55 (0.79/0.42)  \\
\bottomrule
\end{tabular}
\end{table}

\begin{figure}[]
  \centering
  \vspace{-1em}
    \includegraphics[width=0.82\linewidth]{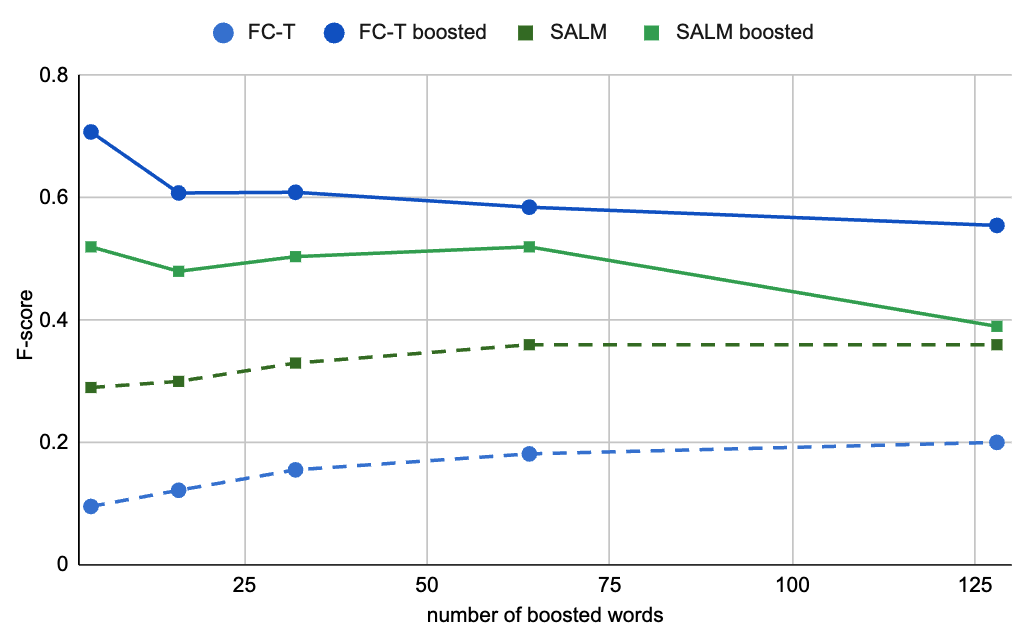}
    \vspace{-1em}
    \caption{Scalability of \# of Boosted Words Comparing Baseline FC-T and In-context Learning based SALM Keyword Boosting}
    \vspace{-1.5em}
    \label{tab:exp-wb-asr-scaleup}
\end{figure}

Figure~\ref{tab:exp-wb-asr-scaleup} studies the scalability of the in-context learning based keyword boosting method for SALM. When scaling up the number of boosted words, both baseline boosting method and SALM
suffer from worse precision with almost unchanged recall. This behavior is caused by a gain of false accepts associated with an increase in the number of candidate  words.
We believe this problem in SALM can be alleviated  by making LLM better handle the long contexts~\cite{openai2023gpt,anil2023palm}.

We look into the win and loss and different error patterns between SALM and baseline in Table~\ref{tab:example}. Generally, SALM performs better on shorter 
 words, compound words, and text normalization, while baseline boosting on FC-T  performs better on longer words and phrases. Nevertheless, SALM suffers from hallucination and early stopping problems that is seldom seen in the baseline. 
 
\begin{table}[t]
\caption{\label{tab:example} {Comparison of Keyword Boosting with SALM v.s. Baseline Fast Conformer-Transducer (FC-T) w/ boosting (errors in \textcolor{red}{red}).} }
\begin{tabular}{lp{0.03\textwidth}|p{0.17\textwidth}|p{0.19\textwidth}}
\toprule
 &           & FC-T                                                                                                                & SALM                                                                                        \\
 \midrule
 \multicolumn{2}{p{0.08\textwidth}|}{ \ Win words} & nvidia, omniverse, rob-otic, cybersecurity, ...  & gpu, hpc,  cudnn, geforce, nvlink,  healthcare, ... \\
\midrule
\midrule
 & Type          & FC-T Hypothesis                                                                                                               & SALM Hypothesis                                                                                        \\
 \midrule
     & text norm     & \textcolor{red}{g t c} is the \textcolor{red}{g p u} computing developers conference                                                                  & gtc is the gpu computing developers conference                                              \\
     \cmidrule{2-4}
 \textbf{Win}      &  boost    & we're \textcolor{red}{sophor} company                                                                                                & we're a software company                                                                    \\
  \cmidrule{2-4}
        &  boost    & tim is the \textcolor{red}{virtuality} driver                                                                                        & tim is the virtual reality driver                                                           \\

        \midrule
    & hallu-cinate & computer graphics is the driving force of the \textcolor{red}{g p}                                                                   & \textcolor{red}{cyberspace} is the driving force of the \textcolor{red}{gpu1 michelangelo sopieness ...}             \\
    \cmidrule{2-4}
  \textbf{Loss}      & hallu-cinate & \textcolor{red}{{[}del error{]}} ladies and gentlemen                                                                                & okay ladies and gentlemen \textcolor{red}{clap your hands cla cla ...} \\
  \cmidrule{2-4}
        & early stop    & i am even the composer of the music you are hearing \textcolor{red}{i ai} brought to life by vivid deep ... & \textcolor{red}{and} i am even the composer of the music you are hearing \textcolor{red}{um are you one cupom}               \\
        \bottomrule
\end{tabular}
\end{table}

\begin{figure}[H]
  \centering
  \vspace{-0.6em}
    \includegraphics[width=0.85\linewidth]{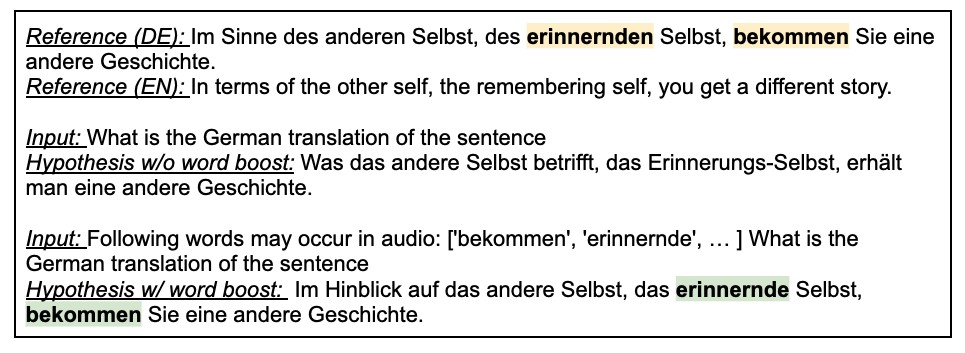}
    \vspace{-1.5em}
    \caption{Example of Using ICL for AST Keyword Boosting.}
    \vspace{-1em}
    \label{fig:wb-ast}
\end{figure}

\vspace{-0.5em}
\subsection{In-context learning for dictionary-guided  translation}
\label{sec:exp-word-boost-ast}
\vspace{-0.4em}

We also conduct  initial studies to see whether above keyword-boosting method can be applied to speech translation in Table~\ref{tab:exp-wb-ast}. We select  40 German words from MuST-C EN-DE dev set with high occurrence in references and low occurrence in hypotheses, and boost them through prompting SALM.

\begin{table}[]
\centering
\vspace{-1em}
\caption{\label{tab:exp-wb-ast} {SALM based AST Keyword Boosting on MuST-C {\footnotesize EN-DE}} }
\begin{tabular}{lcc}
\toprule
systems                               & boost     & F-score ({\em P/R}) \\
\midrule
\multirow{2}{*}{SALM}& N             & 0.20 (0.33/0.15)      \\
                                    & Y             &\textbf{ 0.26} \textbf{(0.25/0.27) }    \\
            \bottomrule
\end{tabular}
\vspace{-1.2em}
\end{table}

Although the overall improvement on F-score is moderate, some successful examples (e.g., Figure~\ref{fig:wb-ast}), show that SALM can correctly pick up the boosted words and the resultant translation is natural. The in-context learning based SALM provides a new route towards  dictionary-guided translation task, where  users want
to guide translation using pre-defined dictionary entries in  inference time~\cite{alkhouli2018alignment}.

\vspace{-0.5em}
\section{Conclusion}
\label{sec:conclude}
\vspace{-0.4em}
We have described {\em SALM}, which
prompts Megatron LLM\cite{shoeybi2019megatron} using NeMo\cite{kuchaiev2019nemo} speech models. We advance  recent Speech-LLM works in two dimensions:  i) utilize the {\em multitask} ability of LLMs to construct a unified model for various speech tasks, as demonstrated by performance on par with bespoke ASR and AST baselines.
 ii) augment speech models with the {\em in-context learning} (ICL) ability of LLMs. We define and study  the ICL of speech-to-text models, and further improve it with {\em speech supervised in-context training}.
We also open-source our implementation to accelerate this line of research.
Future plans include solving the demonstrated hallucination, deletion and long context issues in LLM based SALM.  


\vfill\pagebreak


\bibliographystyle{IEEEbib}
\bibliography{refs}

\end{document}